\documentclass[preprint,12pt]{elsarticle}
\usepackage{graphicx}
\usepackage{amssymb}
\usepackage{lineno}
\usepackage{lineno}
\usepackage{amsmath}
\usepackage{algorithm}
\usepackage{algorithmic}
\usepackage{subfig}
\usepackage{verbatim}
\usepackage[margin=0.7 in]{geometry}
\usepackage{amsmath,mathtools}
\begin{document}
\begin{frontmatter}
\title{A review of UAV Visual Detection and Tracking Methods }
\author[1]{Raed Abu Zitar}
\author[2]{Mohammad Al-Betar}
\author[3]{Mohamad Ryalat}
\author[4]{Sofian Kassaymeh}
\address[1]{Sorbonne Center of Artificial Intelligence, Sorbonne University-Abu Dhabi, Abu Dhabi, UAE.}
\address[2]{Artificial Intelligence Research Center (AIRC), Ajman University, Ajman, United Arab Emirates}
\address[3]{AlBalqa Applied University, Salt, Jordan}
\address[4]{Artificial Intelligence Research Center (AIRC), College of Engineering and Information Technology, Ajman University, Ajman, United Arab Emirates}

\cortext[cor1]{Corresponding Author:raed.zitar@sorbonne.ae}


\begin{abstract}
This paper presents a review for techniques used
for the detection and tracking of UAV's or drones. There are different techniques that depends on collecting measurements of the position, velocity, and image of the UAV and then using them in detection and tracking. Hybrid detection techniques are also presented. The paper is a quick reference for wide spectrum of methods that are used in drones detection process.   
\end{abstract}


\end{frontmatter}
\section{Introduction and literature review}\label{sec:LiteratureReview}

	The process of drone detection itself uses exploring the inter features of flying drones or Unmanned Air Vehicles (UAV). To communicate with the remote operator, drones usually produce radio frequency, heat, and sound signals. Sensor data is gathered by the detection system to validate the existence of drones in the vicinity. It can determine the drones' predicted positions based on the measure \cite{park2021survey}.
	
	UAV detection strategies are listed in the following table based on sensor technology. The subsections that follow examine each detection approach, as well as the underlying mechanism and technological restrictions.
	
	\begin{table}[ht]\centering\scriptsize
	\caption{\label{tbl:DatasetsDetails} UAV Detection Technologies}
	\begin{tabular}{lllll}\hline
		Feature			&  Sensing Devices		&  Advantages							&  Disadvantages						&  Detection Range	\\\hline
		Heat			&  Infrared Camera		&  - Less affected by weather			&  - Low accuracy						&  1-15 km\\
						&  						&  - Long range							&  										&  \\\hline
		RF Signal		&  RF Receiver			&  - Obstacle-free						&  - Unable to detect					&  3-50km\\
						&  						&  - Detect the drone operator			&  - Autonomous fight					&  \\\hline
		Physical Object	&  Radar				&  - Less affected by weather			&  - High expense						&  1-20 km\\
						&  						&  - Long range							&  - Regulations on RF license			&  \\
						&  						&  										&  - Vulnerable to obstacle				&  \\\hline		
		Visibility		&  Optical Camera		&  - Low expense						&  - High affected by weather			&  0.5-3 km\\
						&  						&  - Miniaturized						&  - Vulnerable to obstacle				&  \\
						&  						&  - Identification						&  										&  \\\hline
		Acoustic Signal	&  Acoustic Receiver	&  - Compatible with 					&  - Extremely low detection			&  $<$ 0.2 km\\
						&  						&  RF based sensors						&    range								&  \\
						&  						&  - Miniaturized						&  - Low accuracy						&  \\
						&  						&  										&  - High signal detection				&  \\
						&  						&  										&   complexity							&  \\\hline
	\end{tabular}
	\end{table}

\subsection{Thermal Detection}

	Thermal cameras can recognize and detect substantial amounts of heat emitted by physical components of UAVs, such as batteries, motors, in addition to internal devices \cite{guvenc2018detection}. Detecting drones using their heat signatures has been proposed in several studies. \citet{andravsi2017night} developed a UAV detection technique by detecting thermal power released during flight. To improve the detection system's act and effectiveness and properly recognize drones using their thermal photos, \citet{wang2019towards} used a convolutional neural network (CNN). \cite{HGH2020} system developed by HGH Infrared Systems has the ability to detect infrared radiated from the heat of objects, can surveillance up to 365 degrees.
	

\subsection{RF Scanner}

	Drones that are commanded through operators often communicate specialized data in the form of RF signals that carry flying orders, the output of the sensor,  and other information. The RF scanners gather UAVs' wireless signals and assess whether they are present in the desired region. The basic concepts for RF-based UAVs detection are communication intelligence (COMINT) in addition to Signal intelligence (SIGINT) \cite{park2021survey}. Even though the precision of classification degrades as the number of UAV types (classes number) grows, the accuracy of detection remains satisfactory. For instance, Da-Jing Innovations company produced the Aeroscope system, which is a detection schema that gathers data of UAVs control in a desired area.
	
	

\section{Radar Based Detection}

	The detection Radar process is based on detecting objects and using reflected radio signals to calculate and identify their direction, velocity, range, and form. Radar, unlike the RF scanner, it measures the reflected wave's time-of-flight, while the RF scanner de-modulates the wave. The continuous wave radar calculates the target object's speed by employing distance and data of Doppler. To measure range and speed, coherent pulse-Doppler radar, in addition to Frequency modulated continuous wave (FMCW) radar, records, and tracks sent and received wave phases. Usually, Radar tracking and surveillance utilize a number of bands of frequency \cite{barton2004radar, kouemou2010radar}, that summarize as follow: 

	\begin{itemize}
		\item  The K, Ku, and Ka bands (above 18 GHz), which is short wavelength. Except for marine navigation radar schemas, it was employed for airborne radar systems in their infancy.
		
		\item  The X-band, (8-12 GHz). Synthetic aperture radar is widely used for military reconnaissance systems.
			
		\item   The C-band, (4-8 GHz). Many spaceborne schema such as ERS-1 and 2 and RADARSAT \cite{raney1991radarsat}, and airborne research schema like CCRS Convair-580 and NASA Air-SAR \cite{lee2017polarimetric}, use this technology.
			
		\item  The S-band, (2-4 GHz). weather and Russian ALMAZ satellites radars use it.
			
		\item   The L-band, (1-2 GHz). NASA airborne systems, as well as the US SEASAT and Japanese JERS-1 satellites use it.
			
		\item   The P-band, (300kHz - 1GHz). NASA's experimental aerial research equipment employ the longest radar wavelengths.
	\end{itemize}


	Despite the widespread use of radar in civil and military monitoring schema \cite{park2021survey}, the earliest UAV detection techniques were dubious of employing radar due to exceptionally low UAV RCS \cite{pieraccini2019ground}. A multi-channel passive bistatic radar (PBR) was developed by \cite{liu2017digital} to enhance radar detection resolution, as well as expanded Kalman filter (EKF) and global nearest neighbor (GNN) techniques for adjusting the UAV's position. Numerous UAV detection research shows high granularity FMCW radar with different enhancements such as functional modes, phase interferometry, and other frequencies \cite{drozdowicz201635, jian2018drone, ochodnicky2017drone}. 
	
	Compared to RF scanners, radar -based drone detection has a more significant detection distance and continual observability; nonetheless, there is specific availability of detection and legal constraints. If a UAV hovers in one location or fly at a slow speed, radar cannot differentiate it from any other objects. As a result, integrating radar with other technologies such as RF scanner, cameras is highly advised. Radars continually radiate high-energy RF waves; hence frequency ranges and installation sites need government approval. Due to RF interference difficulties, facilities that currently run radars, such as airports, may have trouble adding more radars \cite{de2021feasibility}.
	
	\cite{Jahangir2016Robust} employ the Holographic radars in order to detect and classify drones with $1 m^2$ size at $20nmi$ range. For this purpose they employ CNN network because it has the ability to reach 98.9\% classification accuracy.
	
	\cite{nemer2021rf} propose a model that use hierarchical classification to detect drone availability. They developed three methods (i.e., Parrot AR, Parrot Bebop, and DJI Phantom). Where the Parrot system, it can detect drone flight mode.

\section{Optical Camera Detection}

	Likewise to thermal cameras, optical cameras have been extensively researched to detect and combat UAVs. \cite{sapkota2016vision} used histograms of oriented gradient characteristics to recognize drones in collected photos. In contrast, \cite{jung2018avss} presented a video-based UAV monitoring schema for surveillance of vast 3-D search space in real-time. Technology-based on optical cameras for detecting UAV is incredibly cheap in expense and has fewer regulations constraints than ones discussed previously, allowing for acceptable tracking schema through the widespread. Nevertheless, limitations like low ranges, significant weather dependence, and obstacle impermeability need to be integrated with other sensor applications. The adoption of military electro-optical/infrared applications for UAV detection is widely used by combining infrared sensors and optical cameras \cite{park2012development}.
	
	
	Deep Convolutional Neural Networks (DCNNs) have become in recent years a cornerstone of the development of visual systems defecated for object detection and tracking (\cite{zhao2019object,soleimanitaleb2019object}). The methods based on learning depend on the principle of deriving feature maps from the input data in the form of images to develop a probabilistic distribution of a set of categories or variables for real values. DCNNs are employed to construct a parametric technique of bounding boxes containing the objects in the context of object detection (\cite{ren2015faster,liu2016ssd,redmon2018yolov3}). However, additional modern techniques employ encoder-decoder architectures for similar purposes (\cite{carion2020end}). Objects have to be allocated an additional identification feature while working within the tracking environment. Following that, the tracking job entails identifying the same occurrence at various temporal dimensions. The majority of object tracking systems designs are established on the tracking-by-detection concept, which involves correlating expected items from conventional detection designs over temporal dimension (\cite{ciaparrone2020deep}).

	
	In (\citep{pawelczyk2020real}), a novel object detection dataset, created for computer vision systems of object detection based on machine learning (ML) methods, is proposed to perform a binary object detection that facilitates automated camera detection of multiple drones. The new dataset extends available datasets (i.e., anti-UAV, MS-COCO, VOC, ImageNet, PASCAL) dedicated to multi-class image classification in addition to object detection through providing a more diverse dataset of images of drones. Real-world footage was used to produce a customized collection of 56 thousand pictures and 55 thousand bounding boxes, then converted into images and hand-labeled to maximize the model's efficacy. Later, the dataset was split into two parts, one for training and the other for testing, and employed to produce six hundred easily deployable Haar Cascades in addition to eight hundred Deep Neural Networks-based models with high performance. Through employing machine learning (ML) methods, they used the dataset to examine different approaches to object detection in order to define a long-term UAV detection system feasibility. Results prove that the Haar Cascade has the ability to be utilized as a Minimum Viable Product model for average performance. At the same time, it failed when compared to Deep Neural Network for a larger dataset.   
	

	In \cite{akyon2021track} owns first place in Drone vs. Bird Challenge organized by AVSS 2021. They address the detection issue by calibrating the YOLO version five model based on synthetic and real data through the use of the Kalman Object Tracker in order to increase detection confidence. Outcomes prove that the performance can be raised through augmenting the synthetic and real data. Furthermore, the temporal data acquired by Kalman Object Tracker can improve performance even more. In addition to use of augmenting data and Kalman Object Tracker, they introduce track boosting technique in order to enhance the detection confidence score. Kalman tracking process utilize velocity and position as base for detection mechanism. Therefore, for tracking, the parameters of tracking were optimized in order to track UAV. Results prove YOLO version five can has reasonable performance for UAV detection by fine-tuned only on synthetically generated data. Furthermore, results show that it is better to combine synthetic data that just use real and synthetic data.  
	
	In (\citep{meng2020unmanned}) proposed three classification models based on convolutional neural network (CNN) are compared experimentally using the transfer learning approach (i.e., VGG16, Inception v3, and ResNet 101), in addition to a couple of original CNN detection models (i.e., SSD, Faster RCNN). Lastly, the obtained UAV test dataset is subjected to an experimental assessment. The image classification method that based on transfer learning employed in this article has significantly improved the results accuracy, recall, and precision when compared to the classic recognition method. Through applying deep learning methods to small samples, the image recognition approach, which is based on transfer learning, is an outstanding way to enhance the accuracy of recognition. To obtain satisfactory outcomes in UAV image recognition, transfer learning transfers the weights of a pre-trained deep neural network and utilizes just tiny sample data. Results show that Inception V3 get the best effect followed by ResNet 101. In addition, results show that Faster R-CNN got better SSD network in detection effect, but SSD network got better detection speed than Faster R-CNN.
	
	In (\citep{liu2020uav}), they design a new technique for object detection based on the YOLO  version 3 model. This technique concentrates on the detection of tiny objects. Also, they gathered a dataset for UAV view to enhance the proposed technique performance, in addition to improve the YOLO model by increasing the field of reception. In order to address the problem of small object miss-detection, they firstly optimize Resblock through joining two ResNet units of the same height and width together in darknet. Secondly, To enrich spatial information, the entire structure of darknet is enhanced through boosting convolution operations at the early layer. Compared with other methods, their results show that the proposed method performs well in a variety of difficulties, particularly in detecting small objects.
	
	\cite{nalamati2019drone} investigate different object detection methods using deep learning models on the training dataset, like Single Shot Detector (SSD), and ResNet-101 and Inception with Faster-RCNN. The main target behind this investigation is tackling the problem of small drones detection in surveillance videos. They conduct their experiments on images extracted from videos. Next, the draw bounding boxes are estimated on the validation dataset, while the IoU and mAP are assisted among estimated bounding boxes against the ground truth. Because of data availability limitation, pre-trained methods were employed to train the CNNs through utilizing transfer learning. The Faster-RCNN model based ResNet-101 architecture obtained the best results in experiments on both training and testing dataset. This research did not focus on measuring the time it takes to detection.

	\cite{wisniewski2021drone} utilize the CNN network through train it on synthetic dataset in order to estimate the drone model in a real-life video feed. The created dataset constructed through applying randomization to orientations, positions, textures, and lighting conditions to the 3D models of the drones. In addition, they applied Gaussian noise to the dataset of training in order to raise classifier performance. Through the investigation, they concluded that fixing the number of hidden layers harms the performance of the classifier in terms of accuracy. They proved that their approach has outperformed other methods, because it is based on drone 3D model. They argue that their method reduced training time in training additional drone models.
	
	
	\cite{seidaliyeva2020real} tackle the issue of real-time UAV detection with reasonable accuracy. The detection process was divided into two steps, detection and classification. In the detection step, the system detect all types moving objects, where the detection is based on background subtraction. In the classification step the system classify objects into three types, drone, bird, background, where classification is done using convolutional neural network (CNN). Results prove that the proposed system has a considerable accuracy comparing the other existing systems at high processing speed. The main limitation of this approach is that it relies heavily on the availability of an moving background.

	\cite{lai2020detection} proposes a framework for moving drones detection based on deep learning in order to predict the distance for the purpose to carry out a feasibility analysis of sense and avoid (SAA) and UAV collision avoidance in mid-air. A monocular camera was used as the sensor for detecting moving objects, in addition to application of deep neural network (DNN) and convolutional neural network (CNN) for the purpose of predicting distance among the invader and the privately owned UAV. The object detection approch used is based on YOLO detector. In addition, the deep neural network and CNN network techniques are employed to evaluate their performance in estimating the distance between moving objects. Also, the VGG-16 approch is employed to extract features from fixed-wing drones, after that, the result is forwarded to the distance network in order to forcast object distance. Using a synthetic images the proposed model was trained, as well as validated using  synthetic and real flight videos. The results reveal that the suggested active vision-based approach is capable of accurately detecting and tracking a moving UAV with low distance errors. 

	\cite{lee2018drone} utilize around two thousand internet images as training dataset for CNN network train phase in order to develop a UAV detection framework  based on ML techniques. This technology is made to work with drones that have cameras. Based on machine classification, the algorithm determines the drone position on the camera photos and the drone vendor model. The proposed model was developed using OpenCV library.

	\cite{behera2020drone} proposed a new drone detection and classification method by employing YOLO version 3 and a convolutional neural network (CNN) with different modalities. The YOLOv3 was used specifically for moving object detection, while CNN was used to accurately extract drone features from images. A convolutional neural network combined with contemporary object detection technologies demonstrates an outstanding way for real-time drone detection.

	\cite{seidaliyeva2020detection} introduced an approach to identify whether the drone is loaded or not by employing YOLO version 2 for UAV image detection. In addition, the proposed framework's preprocessing step conducts the augmentation method to address the data shortage problem. The recent UAV detection research has focused mainly on detecting the presence of UAVs. This is the first time YOLO version 2 has been introduced to detect loaded and unloaded UAV objects using visual data. The dataset was gathered by photographing a multipurpose quad-rotor system in flight.

	For UAV detection and tracking using images obtained by another UAV, \cite{boirel2021deep} present the use of techniques proposed in \cite{akhloufi2019drones, arola2019vision, arola2019uav}. The controls necessary for maneuvering and tracking are deduced from the detected location in real-time. The study propose employing the YOLO version 4 and version 4 tiny together with "Search Area" proposal method for the purpose of UAV detection. The proposed framework was get stimulated through utilizing Unity game engine \cite{Unity2021} and AirSim \cite{shah2018airsim, AirSim2021}.
	
	In (\cite{baykara2017real}), they develop a new framework for the purpose of detection, tracking, and classification for moving vehicles and humans. Their contribution is proposing a framework that works in real-time for low-altitude aerial surveillance. In addition, they utilize the UAV telemetry data in order to compute the classified objects GPS coordinates.
	
	In \cite{haider2019comprehensive}, they propose a schema for detection, tracking, and classification for objects, in addition, to computing detected objects coordinates and velocities. The deep convolution neural network (CNN) distinguishes between vehicles and humans. Furthermore, to detect the object size using an adaptive threshold. The reason behind using deep CNN is its excellent image classification performance. 
	
\section{Acoustic Signal Detection}

	In \cite{sedunov2019stevens}, propose a comprehensive characterization of the developed Drone Acoustic Detection framework by Stevens Institute of Technology, in addition to illustrate the conducted investigations outcomes of UAV acoustic detection under different directional microphones and acoustic arrays.

	

	\cite{casabianca2021acoustic} investigated the CNN, CRNN, or RNN models architecture to identify which one is the model architecture for UAVs acoustic identification in order to apply the network ensemble late fusion later for drones acoustically detection. In particular, using the acoustic signals inputs, they demonstrate the feasibility of using deep neural networks to identify multirotor UAVs, investigate the most suitable model architecture between CNN, RNN, or CRNN is optimal to address the acoustic drone detection and assess the late fusion networks performance against solo models in addition to the selection of the best appropriate voting procedure \cite{dalbah2021modified}, \cite{alomari2021gene}.	

\section{Hybrid Detection Systems}

	Employing only one detection approach leads to a drone detection blind region, making it impossible to eliminate illicit drones properly. 
	The majority of manufacturers use sensor fusion technology and joint hardware control to deploy hybrid drone detection systems \citet{maria2007emotional}, \cite{al2018probability}, \cite{alkoffash2021non}, \cite{zitar2021intensive}. We look at a few examples of their hybrid systems.
	
	

\subsubsection{Radar/ Vision Techniques}

	Drone detection is greatly aided by the use of both radar and optical cameras. By manipulating image tilt, focus, and zoom, vision-based detection can readily track UAVs, but it suffers with dynamic control over the target region. Radar detection, on the other hand, allows for omnidirectional wide-area scanning with low drone identification and scan frequency. As a result, the radar scans the target region, while the vision system regulates the exterior and internal camera parameters to thoroughly probe suspicious locations. As a result, numerous suppliers use this structure since it dynamically adjusts for each other's shortcomings \cite{ELI4030, APS2020, AARTOS2020, DroneShied2020}. 

\section{Conclusion and Possible Future Work}\label{sec:Conclusion}

The advancement in drones detection and classification had been great in the last few years. All possible AI based techniques are investigated including sensors that use images, videos, acoustics,  and RF signals \cite{abu2002performance}, \cite{zitar2005optimum}, \cite{al2018probability}, \cite{al2019kappa}. Deep learning are one of those most successful methods. Other techniques that use classical machine learning methods with features extraction as pre-processing can be investigated in the future \cite{zitar2013genetic}, \cite{afaneh2013virus}, \cite{al2010development}.

\bibliographystyle{elsarticle-num-names}
\bibliography{main.bib}

\end{document}